\documentclass[twocolumn,11pt,a4paper]{article}
\usepackage[utf8]{inputenc} 
\usepackage{graphics}       
\usepackage{times}          
\usepackage{amsmath}        
\usepackage{amssymb}        
\usepackage{subfigure}      
\usepackage{graphicx}
\usepackage{algorithm}
\usepackage[noend]{algpseudocode}
\usepackage{hyperref}
\usepackage{placeins}
\usepackage{booktabs}

\def\figref#1{Fig.~\ref{#1}}
\def\tabref#1{Tab.~\ref{#1}}
\def\eqref#1{Eq.~(\ref{#1})}

\newcommand\etal{\emph{et~al.}}
\begin{document}
\global\def\refname{{\normalsize \it References:}}
\baselineskip 12.5pt
%
%
%
\title{\LARGE \bf Geometrical Stem Detection from Image Data \\for Precision Agriculture}

\date{}

\author{\hspace*{-10pt}
\begin{minipage}[t]{6.9in} \normalsize \baselineskip 12.5pt
\centerline{FERDINAND LANGER\renewcommand{\thefootnote}{\fnsymbol{footnote}}\footnote{The two authors contributed equally to the work.}, LEONARD MANDTLER\footnotemark[1]}
\centerline{ANDRES MILIOTO, EMANUELE PALAZZOLO, CYRILL STACHNISS}
\centerline{Institute of Geodesy and Geoinformation}
\centerline{University of Bonn}
\centerline{Nussallee 15, 53115 Bonn}
\centerline{GERMANY}
\centerline{\{ferdinand.langer, mandtler, amilioto, emanuele.palazzolo, cyrill.stachniss\}@uni-bonn.de}
\end{minipage} 
%
%
\\ \\ \hspace*{-10pt}
\begin{minipage}[b]{6.9in} \normalsize
\baselineskip 12.5pt {\it Abstract:}
High efficiency in precision farming depends on accurate tools to perform weed detection and mapping of crops. This allows for precise removal of harmful weeds with a lower amount of pesticides, as well as increase of the harvest's yield by providing the farmer with valuable information.
In this paper, we address the problem of fully automatic stem detection from image data for this purpose.
Our approach runs on mobile agricultural robots taking RGB images.
After processing the images to obtain a vegetation mask, our approach separates each plant into its individual leaves, and later estimates a precise stem position. This allows an upstream mapping algorithm to add the high-resolution stem positions as a semantic aggregate to the global map of the robot, which can be used for weeding and for analyzing crop statistics.
We implemented our approach and thoroughly tested it on three different datasets with vegetation masks and stem position ground truth.
The experiments presented in this paper conclude that our module is able to detect leaves and estimate the stem's position at a rate of 56 Hz on a single CPU. We furthermore provide the software to the community.
\\ [4mm] {\it Key--Words:}
Stem Detection, Sugar Beets, Weeding, Robotics
\end{minipage}
\vspace{-10pt}}

\maketitle
\thispagestyle{empty} \pagestyle{empty}
%
%

\section{Introduction}
\label{sec:intro}

Robots and automation are key elements of precision farming.
Novel robotics solutions can serve as a dual aid to support crop production,
both by autonomously treating the field in the form of weeding and
fertilizing, and by collecting valuable information about the crop status to provide the 
farmer with feedback that allows for accurate long-term planning.

For this, the robot needs to be able to accurately identify individual plants and
their accurate stem positions. This is needed both to remove the weeds by
coordinating with the mechanical or spraying tool mounted on the robot, and to
map each individual crop in the global map in order to be able to track each crop plant
in time, which provides growth statistics to the farmer.

\begin{figure}[t]
\vspace{2mm}
\centering
\setlength\tabcolsep{0.002\linewidth}
\begin{tabular}{cc}
  \includegraphics[width=0.56\linewidth]{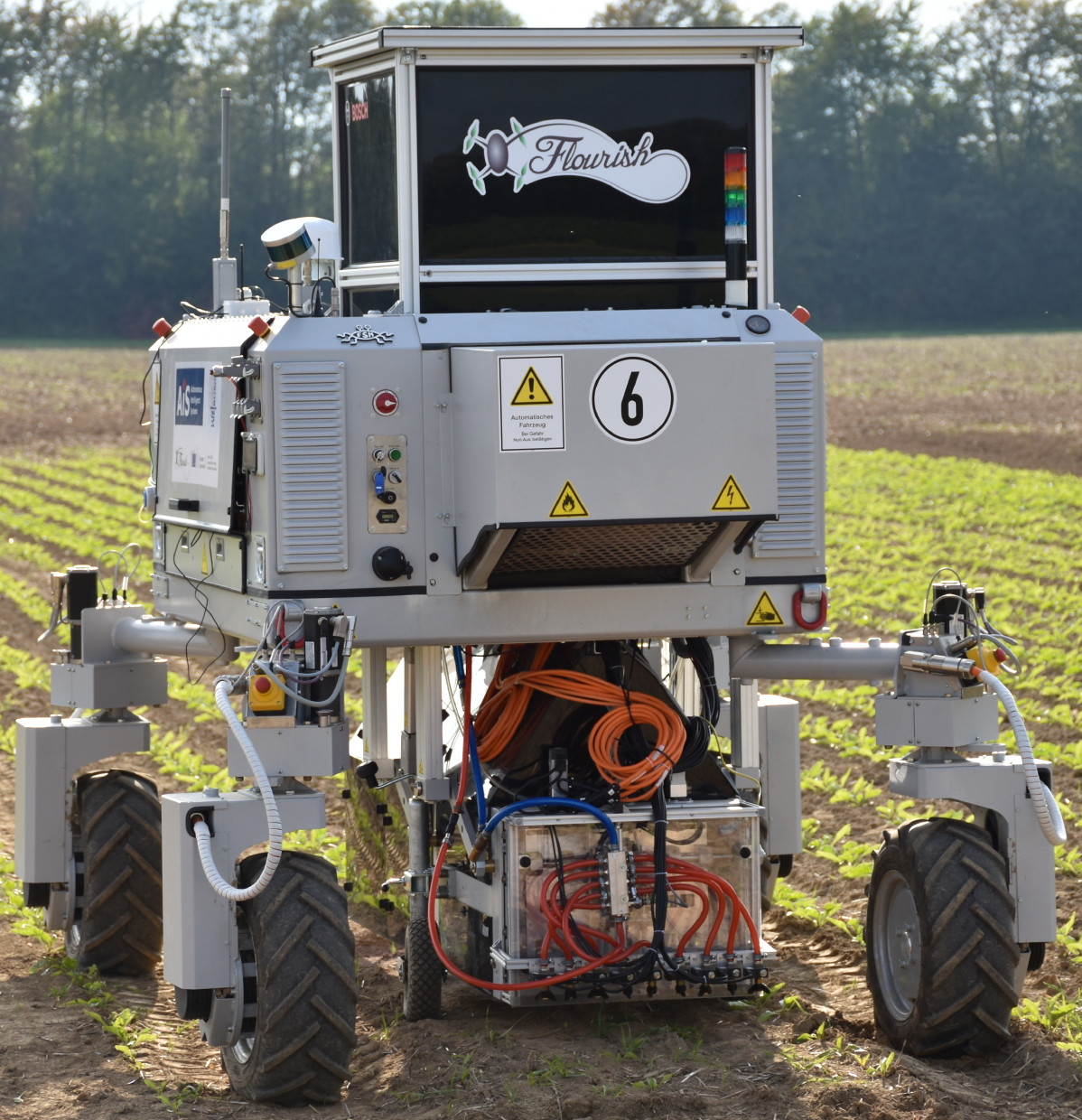} &
  \includegraphics[width=0.435\linewidth]{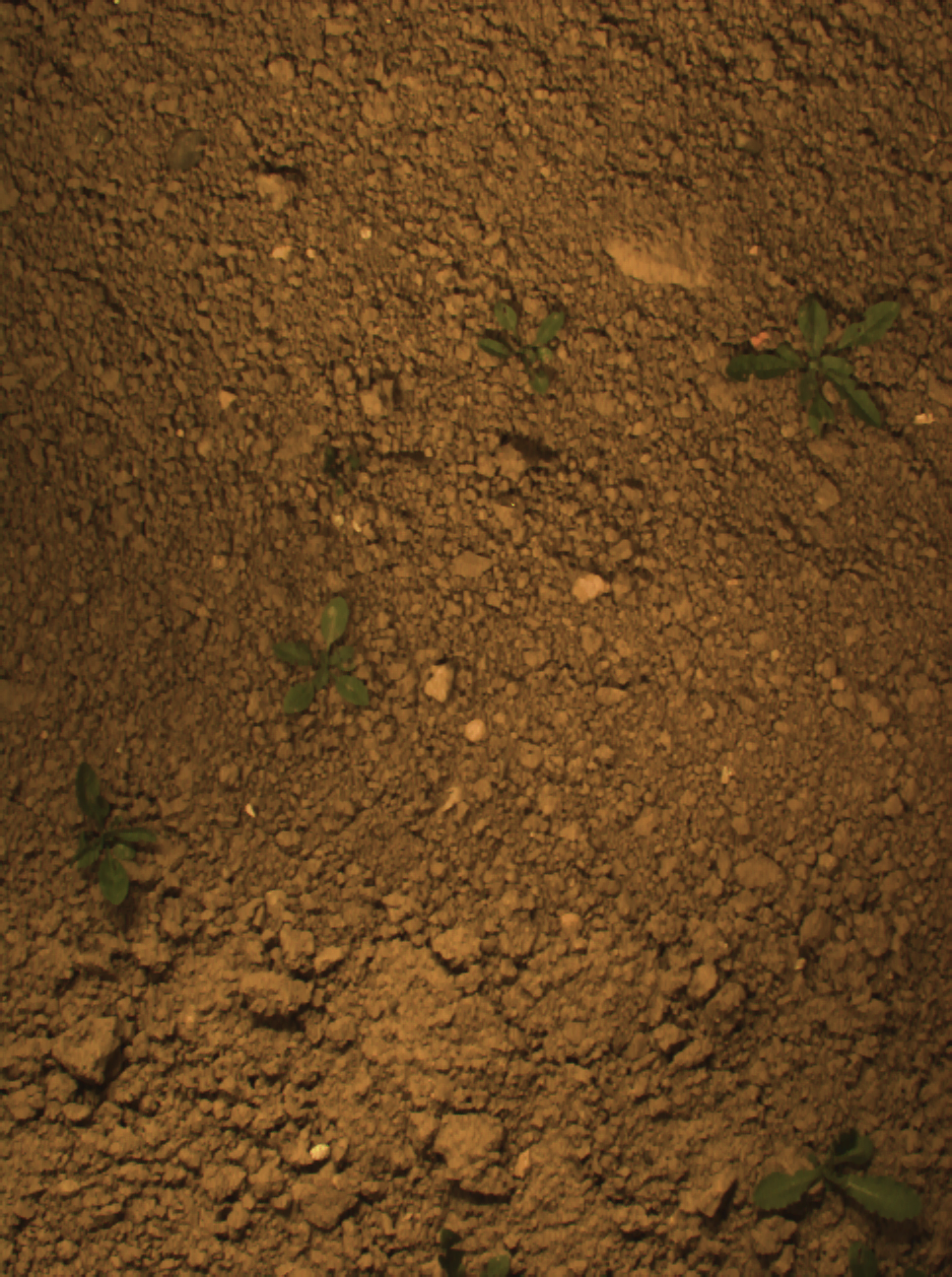}
\end{tabular}
\caption{Left: Agricultural robot selectively stamping or spraying weeds, and 
mapping individual plant growth stages, both made possible by stem detection. Right:
Sample image taken from the robot. Best viewed in color.}
\label{fig:samples_front_page}
\end{figure}

Our approach targets the stem detection problem for autonomous robots from vision 
data, such as the one depicted in \figref{fig:samples_front_page}. 
An uninformed and commonly used
strategy is to use the center of mass of each plants' vegetation area, but as leaves
usually tend to grow unevenly, this is inaccurate in most cases, especially in
late growth stages. For many types of plants we can distinguish several leaves
and we use this prior knowledge in our approach to improve performance.

\begin{figure}[t]
  \centering
 \includegraphics[width=0.99\linewidth]{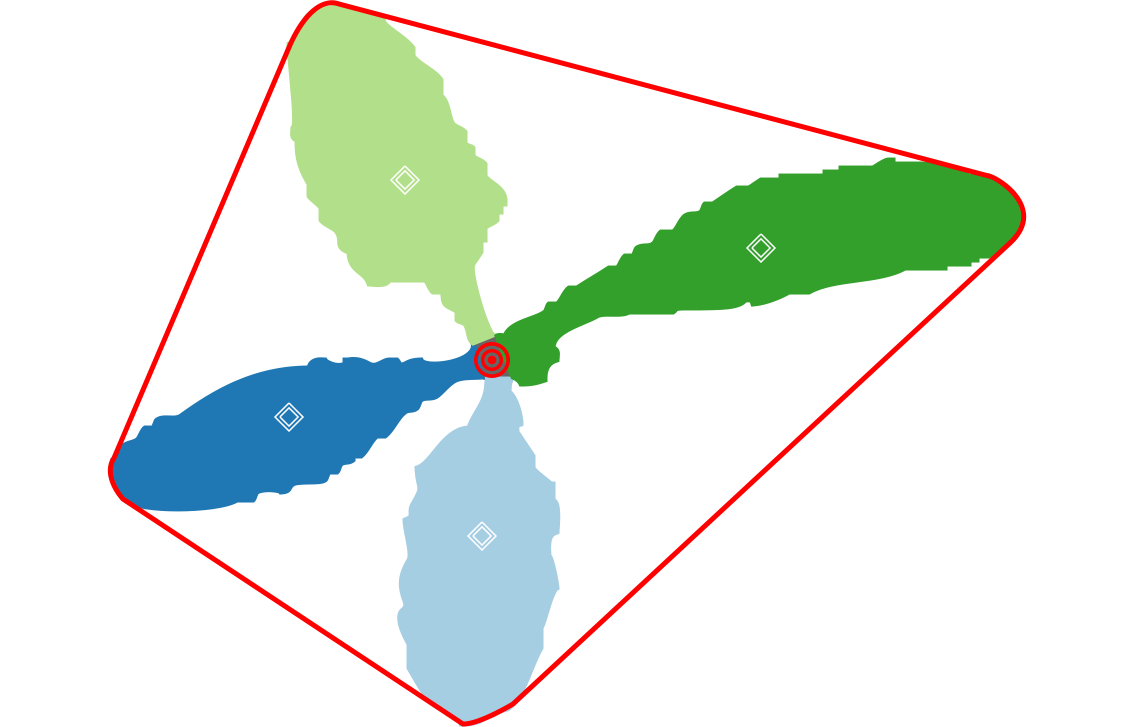}
 \caption{Two-step approach for a geometric stem detection. An initial leaf detection is performed using convexity defects in the convex hull of the vegetation mask. The final stem detection is obtained from connection of leaves.}
  \label{fig:motivation}
\end{figure}

Our stem detection pipeline uses a geometric approach, related to Midtiby~\etal~\cite{midtiby2012biosyseng,midtiby2012icae}.
Our three-step approach, shown in \figref{fig:motivation}, detects leaves in each vegetation object
in an RGB (or RGB plus near infra-red) image and infers the stem as a connection of the leaves.


The main contribution of this paper is a fast visual stem detection tool which
serves as the perception backbone of an agricultural robot that targets both
weeding and crop mapping. Our pipeline is implemented in C++ using the Robot Operating
System (ROS) for sensor and actuator communication and testing, and it can run at over
the frame-rate of a commercial camera.


\newcommand{\ClaimHz}{20\,Hz}

In sum, we make three key claims:
Our approach is able to
(i)~detect leaves by a geometric approach,
(ii)~improve the performance of stem detection compared to the typically used center of mass approaches and
(iii)~process images faster than the frame-rate of a commercial camera.
These claims are backed by our experimental evaluation, and the approach is available
as open-source code for the community to use at \url{https://github.com/Photogrammetry-Robotics-Bonn/geometrical_stem_detection}.

\section{Related Work}
\label{sec:related}

The exploitation of semantics from the environment is an active area of research 
in agriculture technology ~\cite{lottes2016icra,milioto2018icra,mueter2013}.
Robotics has the potential to address the task accurately and efficiently, 
jointly using these semantics for both automatic weeding and mapping of individual
plants to provide accurate yield information to the farmer. Some examples of 
weeding are the work by Mueter~\etal~\cite{mueter2013}, which focuses on the removal of weeds
through the design of a mechanism for intra-row weeding using vision, and the
work by Nieuwenhuizen \cite{nieuwenhuizen09phd}, who presents an approach for
automated detection and control of volunteer potato plants. 
On the crop plant detection side, Kraemer~\etal~\cite{kraemer2017iros} aim at 
using individual crop stems over a temporal period, which not only allows for 
assessing individual crop plant progress, but also serve as landmarks for
localization of the robot in the field.

It is important to note that, even though there has been a significant amount of
research done on plant classification and semantic 
segmentation~\cite{lottes2016icra,lottes2017iros,milioto2017isprsannals,milioto2018icra}
for its use in robotics, only a selected number target the specific case of stem
detection. It is imperative to solve this task for both weeds and crops, in order
to allow for autonomous weeding and long term plant statistics, which is why we focus
on this specific problem, and not on the classification pipeline.

Most visual state-of-the-art methods for the task of stem detection rely on a two step approach. 
First, the images are segmented into background and vegetation areas, which allows the approach
to rely solely on the latter, improving the robustness to different soil conditions, and 
making the underlying algorithms faster by only analyzing relevant parts of the image. Second, the 
vegetation areas are analyzed looking for stems. 

Different approaches to the vegetation segmentation task have been proposed 
in the field~\cite{Guo2013cea,hamuda2016cea,milioto2018icra,TorresSanchez15cea}. 
Guo~\etal~\cite{Guo2013cea} apply a learning approach for vegetation detection using
a decision tree. 
Hamuda~\etal~\cite{hamuda2016cea} survey different plant segmentation 
methods in field images by analyzing several threshold and learning based approaches.
Torres-Sanchez~\etal~\cite{TorresSanchez15cea} investigate an adaptive and automatic
thresholding method for vegetation detection based on the Normalized Difference 
Vegetation Index~(NDVI) and the Excess Green Index~(ExG). They report a detection
rate of around $90-100$\,\%.
Milioto~\etal~\cite{milioto2018icra} propose a convolutional neural network classifier
combining learning with classical vegetation indexes without manual selection of 
hyperparameters. 
In our work, we use an automatic thresholding method over different widely
used vegetation indexes, NDVI and Excess Green, relying on RGB+NIR and RGB-only imagery, respectively. 
Because our work highly depends on a well-performed segmentation,
we perform experiments on different datasets with various accuracies in the vegetation mask.
The code provides an implementation of several vegetation segmentation approaches, as well as an option to provide the masks externally as an input, in order to allow for the implementation of new methods as they
become available.

After the vegetation segmentation is performed, straightforward solutions to infer
the position of each vegetation object's stem like the center of mass have been proposed
in the literature, such as the stem detection part of the work by Kiani and Jafari~\cite{kiani2012jast}.
However, as most plants do not grow in a symmetrical manner this approach is limited,
and therefore it should only be used as a fallback method.
Haug~\etal~\cite{haug2014raar} work around this problem by introducing a sliding window
classifier to predict the position of a stem. The classification map is post-processed
to regress viable stem positions.
Some approaches have also been proposed for leaf segmentation, where researchers have
investigated methods using structural operators and hand-crafted features~\cite{hall2015wacv,hemming05ah,midtiby2012biosyseng,midtiby2012icae,wang08amc}.
Wang~\etal~\cite{wang08amc} segment leaf images by using morphological operators and
shape features and apply a moving center hypersphere classifier to infer the
plant species, instead of the stem position.
Hall~\etal~\cite{hall2015wacv}  evaluate the effectiveness of traditional hand-crafted
features and propose the use of deep CNN features for leaf classification.
Hemming~\etal~\cite{hemming05ah} use a sequence of images and leaf movement to segment the leaves 
and an automatic thresholding to obtain individual leaf position.
Midtiby~\etal~\cite{midtiby2012biosyseng} use a geometric approach in two steps.
First, they extract the leaves of a plant, and then use this information to predict the stem emerging point.
They use the curvature along the plant's contour for leaf segmentation.
In another work by Midtiby~\etal~\cite{midtiby2012icae}, they introduce a convex-hull based leaf detector.
The stem's location is then predicted by using a multivariate normal distribution using the information the leaves provide. Our approach for the stem detection step differs from the original approach by Midtiby~\etal~\cite{midtiby2012biosyseng} in that we take the intersections of the leaf directions as the stem position,
instead of relying on a multivariate normal distribution using all leaf distributions. This makes the
approach simpler and faster to run, but it requires a post-processing of the result to analyze the feasibility
of the regression in order to avoid giving wrong predictions where the approach does not apply.
We discuss the detailed differences in the following section.

\section{Our Approach}
\label{sec:main}
\vspace{-4pt}

The main goal of our approach is to detect in real-time the stem of 
non-overlapping plants in an early growth stage.
Given an RGB image acquired from a mobile robot, we identify the plants
and separate them from the ground using the Excess Green index. 
From the obtained masks, we identify the leaves from the convexity defects of
the plant's shape. Finally, we compute the stem position from the directions of
the leaves.

\begin{figure*}[t]
  \centering
  \subfigure[Mask after closing operation.]{\includegraphics[width=0.22\linewidth]{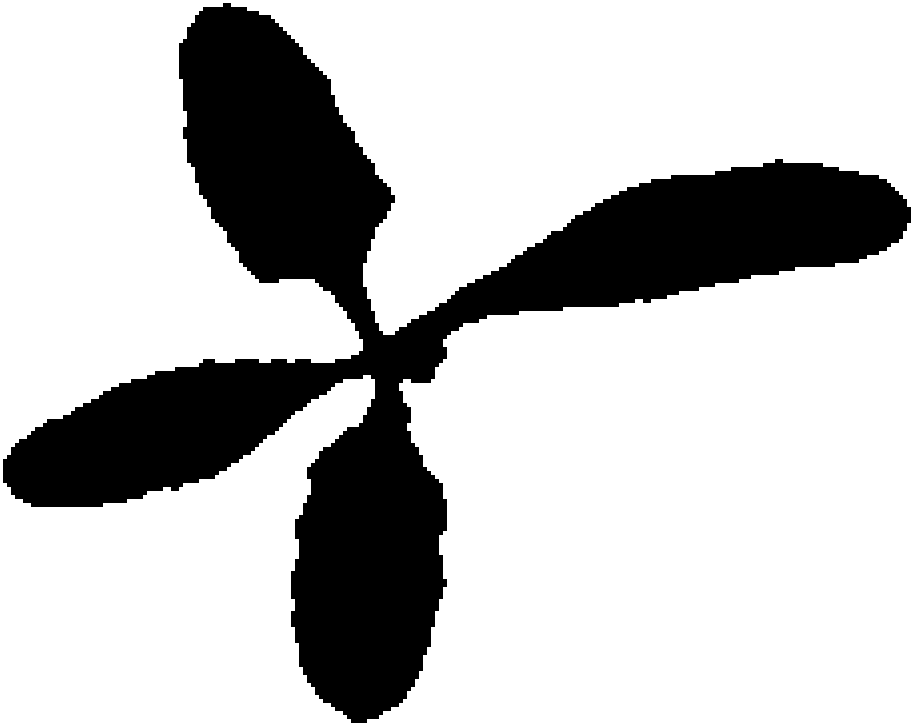}\label{fig:steps_leaf_1}}\hfill
  \subfigure[Contour and convex hull.]{\includegraphics[width=0.22\linewidth]{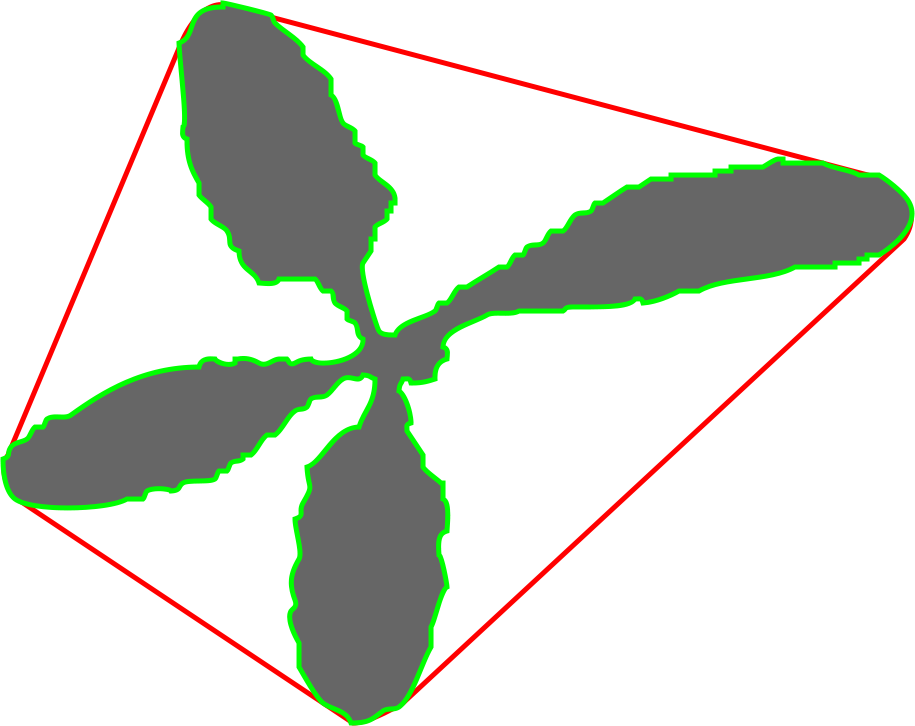}\label{fig:steps_leaf_2}}\hfill
  \subfigure[Cut-off points for leaves.]{\includegraphics[width=0.22\linewidth]{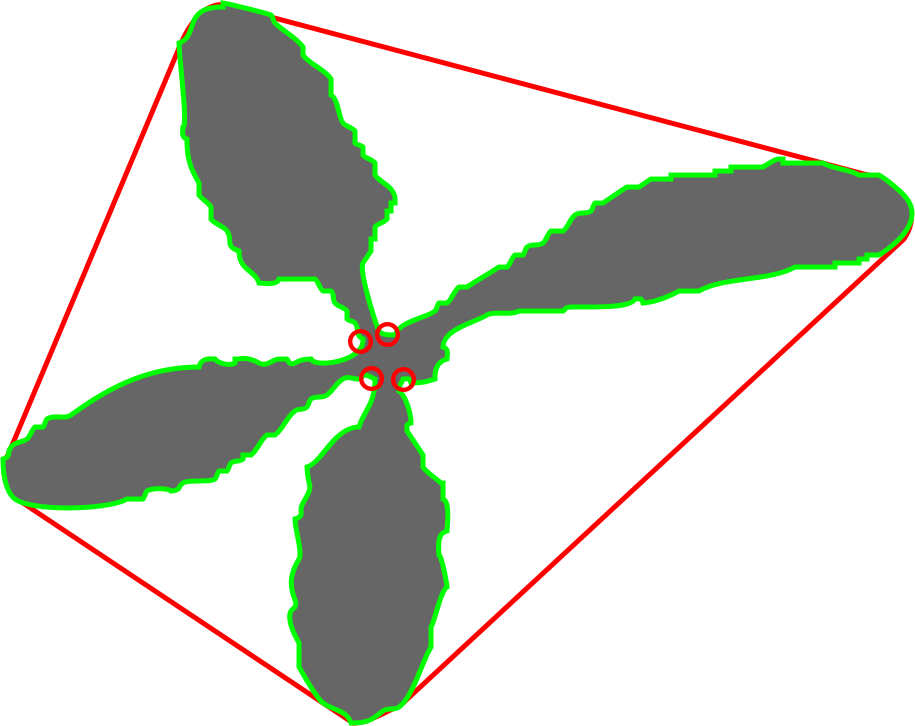}\label{fig:steps_leaf_3}}\hfill
  \subfigure[Extract leaves.]{\includegraphics[width=0.22\linewidth]{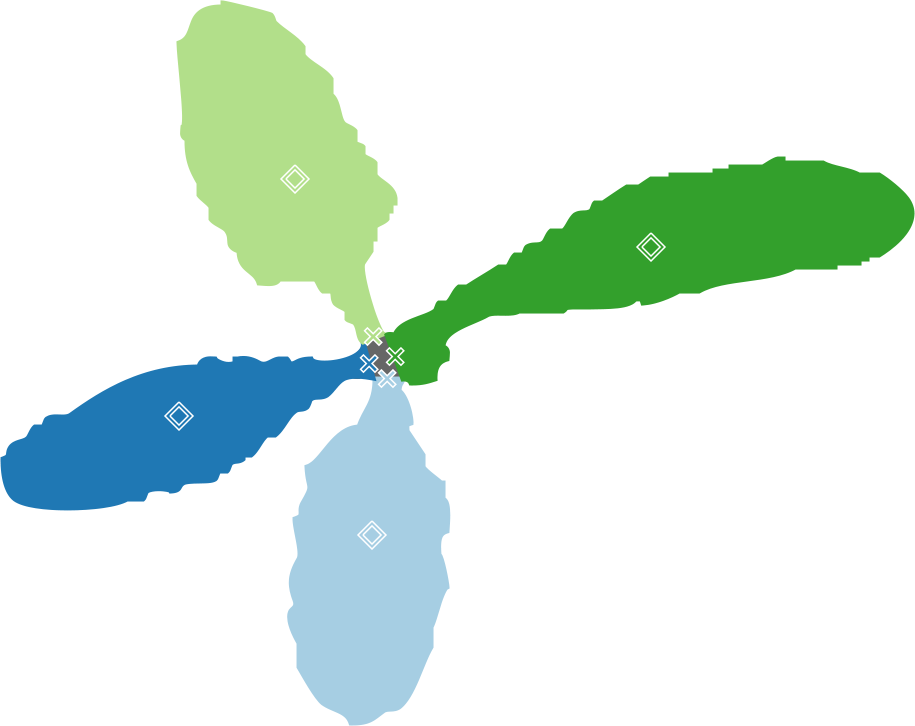}\label{fig:steps_leaf_4}}\hfill
  \caption{Steps of leaf detection which are performed for each mask object. The algorithm provides a list of the individual leaves of a plant. Best viewed in color.}
  \label{fig:steps_leaf}
\end{figure*}

\subsection{Vegetation Segmentation}

The first step needed by the pipeline is the generation of a binary mask separating
the areas of vegetation from the background, as in \eqref{eq:vegetationMask}, in 
order to analyze the former in object space in search for stems

\begin{equation}
\mathcal{I}_\mathcal{V}(i,j) = \left\{\begin{array}{ll} 1, & \text{if~} \mathcal{I}(i,j) \in \mathit{vegetation} \\ 0, &  \text{otherwise}\end{array}\right.,
\label{eq:vegetationMask}
\end{equation}
with the pixel location $(i,j)$. Our pipeline currently supports the generation of
the masks using the NDVI index for RGB+NIR images~(see \eqref{eq:NDVI}) and the Excess
Green, for RGB-only~(see \eqref{eq:ExG}), and a posterior binarization 
using Otsu and Triangle automatic thresholding.

\begin{equation}
\mathcal{I}_{\mathrm{NDVI}}{}(i,j) = \frac{\mathcal{I}_{\mathrm{NIR}}{}(i,j)  - \mathcal{I}_{\mathrm{R}}{}(i,j) }{\mathcal{I}_{\mathrm{NIR}}{}(i,j)  + \mathcal{I}_{\mathrm{R}}{}(i,j) }
  \label{eq:NDVI}
\end{equation}

\begin{equation}
\mathcal{I}_{\mathrm{ExG}}{}(i,j) = 2\times\mathcal{I}_{\mathrm{G}}{}(i,j)  - \mathcal{I}_{\mathrm{R}}{}(i,j) - \mathcal{I}_{\mathrm{B}}{}(i,j)
  \label{eq:ExG}
\end{equation}

The code available also allows to include the masks as inputs, in order to be able
to implement new segmentation methods as they become available in the community.

\subsection{Leaf Detection}
\label{ssec:sec3_ourapproach:leaf_detection}

Once we obtain a binary vegetation mask from the captured image, we proceed to 
identify the leaves of each plant.
We first pre-process each mask by performing a closing operation on the image, to
remove noise and artifacts and to connect masks that incorrectly appear as separate blobs.
The closing operation is done with an elliptic kernel, with an optimal kernel size 
that depends on the resolution of the input image and the quality of the mask,
and it is a hyper-parameter of our approach.

Given that we need to count the leafs for each plant object, and
that there are multiple objects per image, we then proceed to separate each image mask
into segments containing one object each, by analyzing the connected components, which
assumes non-overlapping plant objects. \figref{fig:steps_leaf_1} shows an example of an
object mask after the preprocessing step. 

To identify the leaves, we find the contour (green line in \figref{fig:steps_leaf_2}) of each mask object and then compute its convex hull (red line in \figref{fig:steps_leaf_2}).
The convexity-defects of a mask are the local maximum distances from the contour to the convex hull.
We consider all the convexity-defects that have a distance from the convex hull greater than a predefined threshold, depending on the dataset (red circle in \figref{fig:steps_leaf_3}). This threshold is the
second hyper-parameter of our approach.

Finally, each neighboring pair of convexity-defects defines one leaf by 
representing its cut-off segment. Each leaf is then defined by the pair of 
cut-off points and all the contour points between them.
\figref{fig:steps_leaf_4} shows a plant with four convexity-defects, which 
define four individual leaves.
We compute the center of mass of the leaf, which we define as the \emph{leaf~center} (white diamond in \figref{fig:steps_leaf_4}), and the mean of the corresponding cut-off-points, 
which we define as the \emph{leaf~root} (white cross in \figref{fig:steps_leaf_4}),
for each leaf.
The connecting line of leaf center and leaf root is the \emph{leaf~direction}.

\subsection{Stem Detection}
\label{ssec:sec3_ourapproach:stem_detection}

\begin{figure}[t]
  \centering
  \includegraphics[width=0.7\linewidth]{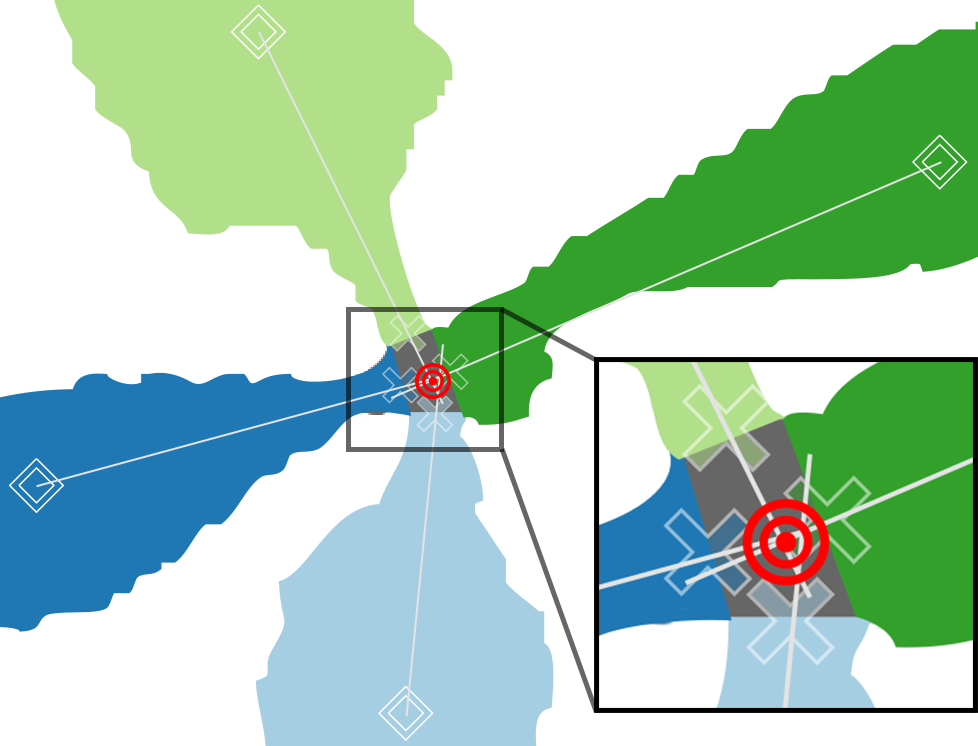}
  \caption{Key features of each leaf: (i) leaf center (white diamond); (ii) leaf root (white cross); (iii) leaf direction (white line). Stem point (red circle) as the intersection of all leaf directions. Best viewed in color.}
  \label{fig:steps_stem}
\end{figure}

\begin{figure*}[]
  \centering
  \subfigure[Dataset A - Stuttgart - late stage]{\frame{\includegraphics[width=0.27\linewidth]{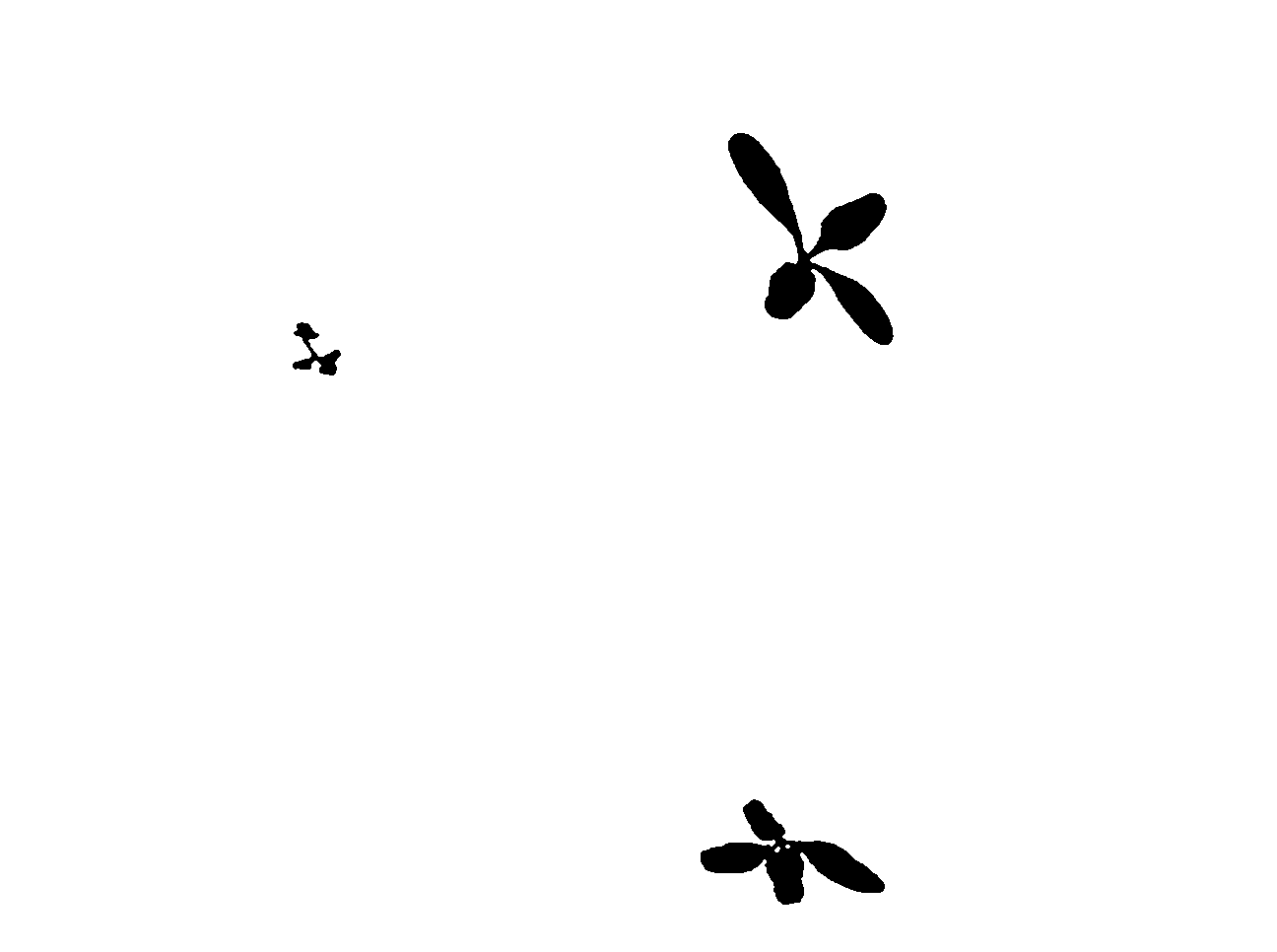}}\label{fig:sample_Renn_150629}}\hfill
  \subfigure[Dataset B - Stuttgart - early stage]{\frame{\includegraphics[width=0.27\linewidth]{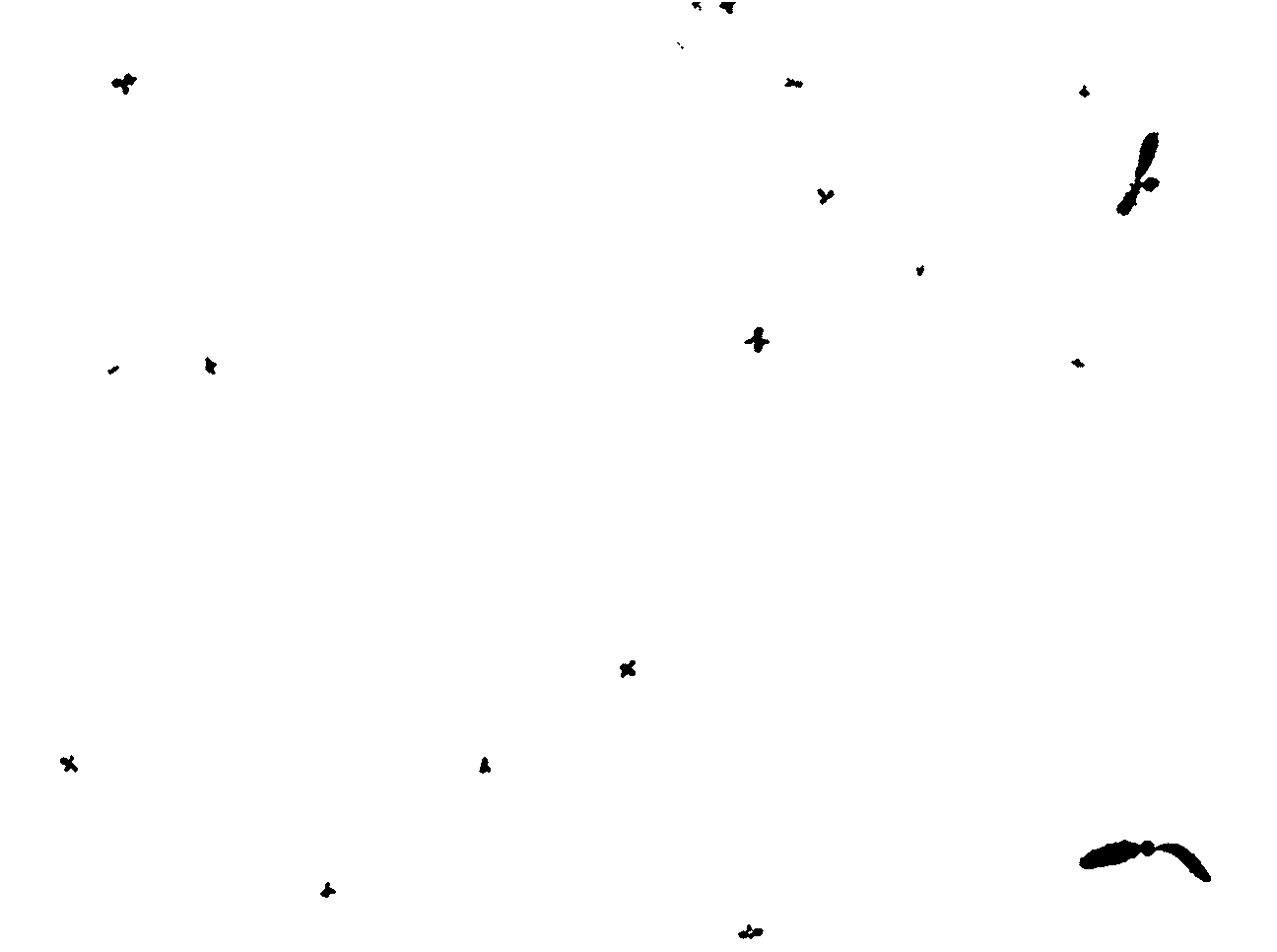}}\label{fig:sample_Renn_150513}}\hfill
  \subfigure[Dataset C - Bonn - late stage]{\frame{\includegraphics[width=0.27\linewidth]{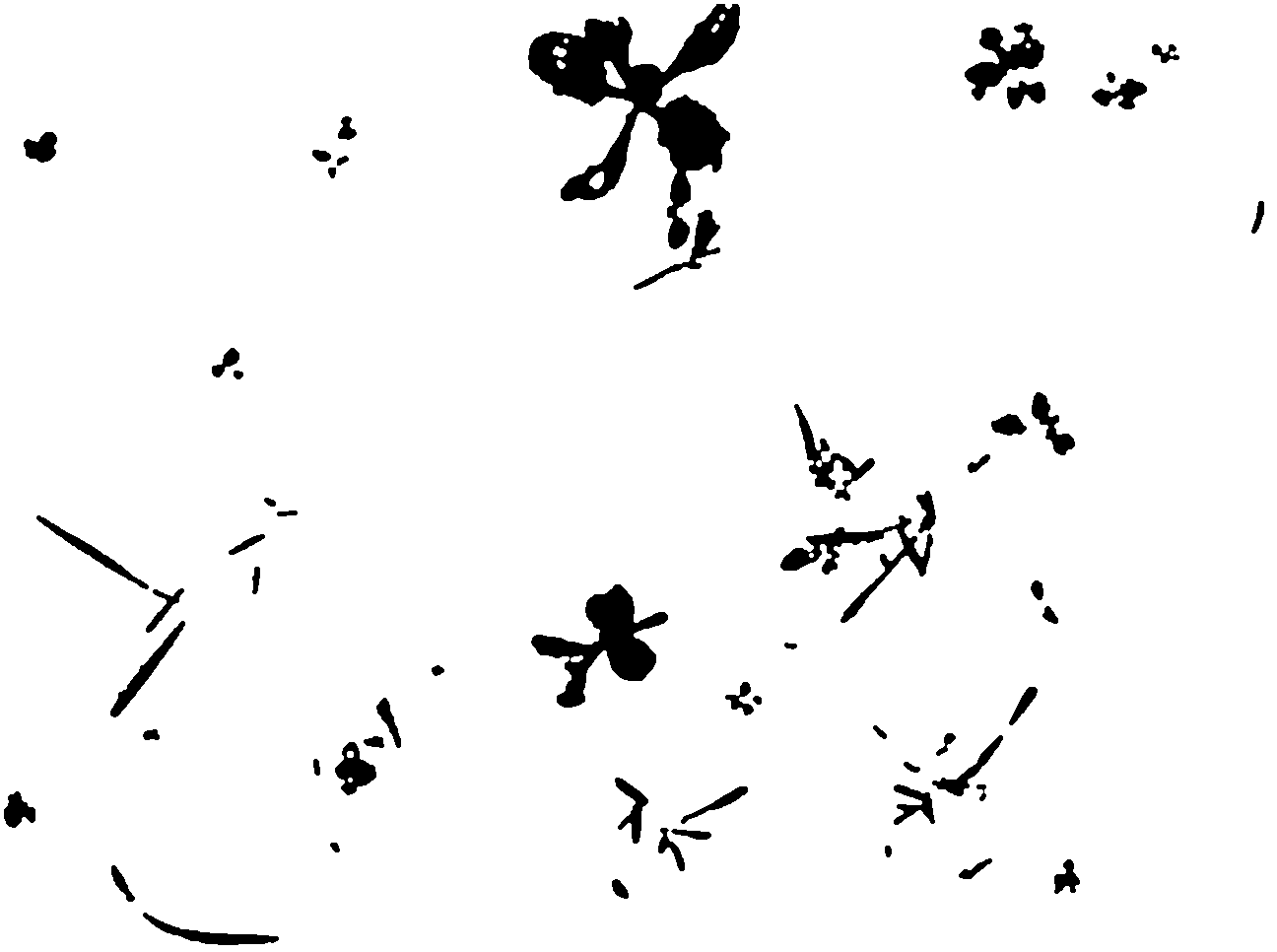}}\label{fig:sample_CKA_160523}}

  \caption{Sample mask images from all three datasets used to evaluate the method.}
  \label{fig:sample_masks}
\end{figure*}

To compute the stem position of a plant we first compute the pairwise intersections
of all the leaf directions. The mean point of the image coordinates of all 
intersections is what we use to regress the position of the stem in the image
for each vegetation object. \figref{fig:steps_stem} illustrates the process described.
In case of less than $N$ detected leaves, where $N$ is a parameter describing the minimum of
leaves present in the currently plant growth stage, the described approach is not
used to estimate the stem's position. In such cases, we directly consider
the center of mass of the mask object as the position of the stem, as a fall-back method.

\section{Experimental Evaluation}
\label{sec:exp}

The main focus of this work is a stem detection method for non-overlapping growth
stages in real time. Our experiments are designed to show the capabilities of 
our approach and to support our key claims, which are:
(i)~to accurately detect leaves by a geometric approach,
(ii)~to improve the performance of stem detection compared to the typically used center of mass approach and
(iii)~to achieve this at a rate higher than the frame rate of a commercial camera.

\subsection{Performance}

\begin{table}[t]
  \centering
  \caption{Results for dataset A}
  {\footnotesize
    \begin{tabular}{ccc}
      \toprule
      approach & recall & precision\\
      \midrule
      center of mass, $m=10$ & 70.5\,\% & 57.9\,\% \\
      our approach, $m=1$ & 91.1\,\% & 63.8\,\% \\
      our approach, $m=6$ & 87.6\,\% & 70.0\,\% \\
      our approach, $m=10$ & 82.2\,\% & 67.6\,\% \\
      our approach, $m=20$ & 75.9\,\% & 63.6\,\% \\
      \bottomrule
    \end{tabular}
    \label{tab:perf_Renn_150629}
  }
  \caption{Results for dataset B}
  \centering
  {\footnotesize
    \begin{tabular}{ccc}
      \toprule
      approach & recall & precision \\
      \midrule
      center of mass, $m=10$ & 89.1\,\% & 67.6\,\% \\
      our approach, $m=1$ & 94.3\,\% & 66.4\,\% \\
      our approach, $m=4$ & 92.9\,\% & 68.9\,\% \\
      our approach, $m=10$ & 90.3\,\% & 68.4\,\% \\
      our approach, $m=20$ & 89.0\,\% & 71.3\,\% \\
      \bottomrule
    \end{tabular}
    \label{tab:perf_Renn_150513}
  }
  \caption{Results for dataset C}
  \centering
  {\footnotesize
    \begin{tabular}{ccc}
      \toprule
      approach & recall & precision \\
      \midrule
      center of mass, $m=10$ & 53.6\,\% & 25.1\,\% \\
      our approach, $m=1$ & 64.8\,\% & 13.5\,\% \\
      our approach, $m=10$ & 58.7\,\% & 27.6\,\% \\
      our approach, $m=20$ & 52.6\,\% & 29.9\,\% \\
      \bottomrule
    \end{tabular}
    \label{tab:perf_CKA_160523}
  }
\end{table}

The first experiment is designed to show the performance of our approach and to support our claims (i) and (ii), i.e., our approach accurately segments individual plant leaves and outperforms the center of mass approach in a variety of settings, from different soil types to different growth stages.
To evaluate our approach, we measure the performance in terms of \emph{precision} and \emph{recall} for different values of the kernel size $m$, for the morphological closing operation.
The \emph{precision} measure represents the number of correctly detected stems per estimated stems.
The \emph{recall} measure describes the ratio of correctly detected stems by the total number of stems present in the annotated ground truth. In order to determine what a proper detection is, we define a threshold for a true positive of 0.5~cm, considering any detection out of this range from the ground truth as a false positive, and any missing detection as a false negative.
We tested and compared both the center of mass approach and our method on three different dataset recorded in Stuttgart and Bonn, in Germany, containing different soil types and growth stages.

Dataset A, from Stuttgart, Germany, includes in each frame a small amount medium-sized sugarbeets and some scattered weeds, see \figref{fig:sample_Renn_150629} for an example.
Dataset B, also from Stuttgart, was taken at an earlier growing stage of the sugarbeets, and the images contain several very small weeds, see \figref{fig:sample_Renn_150513}.
Finally, dataset C, from Bonn, Germany, is substantially different and more challenging compared
to the other two. In particular, it includes more plants, some of which do not have a stem such as grass,
and the obtained vegetation masks are in general more fragmented, requiring a larger
kernel size to close them. See \figref{fig:sample_CKA_160523} for an example.

\tabref{tab:perf_Renn_150629}, \tabref{tab:perf_Renn_150513} and \tabref{tab:perf_CKA_160523}
show the results of our approach on the three dataset.
In the case of dataset A, our approach shows an improvement of $5~\%$ to $21~\%$ for the recall and from $6~\%$ to $13~\%$ for the precision, compared to a standard center of mass approach.
In dataset B, the gain is up to $5~\%$ for the recall and $4~\%$ for the precision. The reduced gain is due to the fact that smaller plants do not have strong leaf characteristics and their resolution is limited.
For dataset C, our approach shows an improvement of $3~\%$ for the recall and $2~\%$ for precision, with
the same kernel size. 

\subsection{Runtime}
\FloatBarrier

The next experiment is designed to support our claim (iii), i.e., that our approach runs
fast enough to support online processing on the robot in real time.

\tabref{tab:speed} summarizes the runtime results for the vegetation segmentation
and our stem detection node.
The numbers support our third claim, namely that the computations can be executed 
fast and in an online fashion.
On a state-of-the-art mobile Intel i7 processor, we obtain an average frame rate of 56\,Hz.
As most robotic weeding tools can operate at 5\,Hz, our approach is fast enough
to be used online on an agricultural robot.

\begin{table}[h]
\caption{Average runtime and std.~dev.}
\centering
{
\begin{tabular}{cc}
\toprule
module & i7-6700HQ CPU 2.5\,GHz\\
\midrule
%
mask & 44.6\,ms~$\pm$~4.3\,ms~$\approx$~22\,Hz \\
%
\textbf{stem} & \textbf{17.7\,ms~$\pm$~6.6\,ms~$\approx$~56\,Hz} \\
%
$\sum$ &  62.2\,ms~$\pm$~10.9\,ms~$\approx$~16\,Hz \\

\bottomrule
\end{tabular}
}
\label{tab:speed}
\end{table}

In summary, our evaluation shows that our method provides improved stem detection performance for vegetation masks with distinct leaf characteristics.
At the same time, our method is fast enough for online processing and has small memory demands.

\section{Conclusion}
\label{sec:conclusion}

In this paper, we presented a geometric fast processing approach to stem detection.
Our approach operates in real time using low resources and only depends on three hyper-parameters,
which are the kernel size of the closing operator, the distance threshold for the convexity defects, and
the number of leaves depending on the growth stage.
Our method exploits the geometric structure of plants by first detecting individual leaves and then estimating the stem's position.
We implemented and evaluated our approach on different datasets and provided comparisons to other existing techniques.
The approach depends on the growth stages of plants as we need them non-overlapping.
However, we expect high improvement against center of mass approaches on large plants with higher potential of asymmetric growth.
The limiting factor for our approach is the provided vegetation mask. Visual noise in the mask leads to weaker performance, and we therefore separate the implementation of the stem detection from the vegetation segmentation,
in order to include newer, more novel, methods as they become available.
%

\bibliographystyle{plain}

\begingroup
    \small{
      \bibliography{glorified,new}
    }
\endgroup








\end{document}